# Argumentative inference in uncertain and inconsistent knowledge bases

Salem Benferhat – Didier Dubois – Henri Prade
*Institut de Recherche en Informatique de Toulouse (I.R.I.T.)*
*Université Paul Sabatier – C.N.R.S.*
*118 route de Narbonne – 31062 TOULOUSE – FRANCE*
*email: {Benferha, Dubois, Prade}@irit.fr*

## Abstract

This paper presents and discusses several methods for reasoning from inconsistent knowledge bases. A so-called argumentative-consequence relation, taking into account the existence of consistent arguments in favor of a conclusion and the absence of consistent arguments in favor of its contrary, is particularly investigated. Flat knowledge bases, i.e. without any priority between their elements, as well as prioritized ones where some elements are considered as more strongly entrenched than others are studied under different consequence relations. Lastly a paraconsistent-like treatment of prioritized knowledge bases is proposed, where both the level of entrenchment and the level of paraconsistency attached to a formula are propagated. The priority levels are handled in the framework of possibility theory.

## 1. Introduction

One of the emerging important problems pertaining to the management of knowledge-based systems is inconsistency handling. Inconsistency may be present for several reasons: the presence of general rules with exceptions, the existence of several possibly disagreeing sources feeding the knowledge base are among the most common ones. There are two attitudes in front of inconsistent knowledge. One is to revise the knowledge base and restore consistency. The other is to cope with inconsistency. The first approach meets two difficulties: there are several ways of restoring inconsistency yielding different results, and the problem is that part of the information is thrown away and we no longer have access to it. Coping with inconsistency bypasses these difficulties. However we must take a step beyond classical logic, since the presence of inconsistency enables anything to be entailed from a set of formulas.

This paper investigates several methods for coping with inconsistency by suitable notions of consequence capable of inferring non-trivial conclusions from an inconsistent knowledge base. These consequence relationships coincide with the classical definition when the knowledge base is consistent. When the knowledge base is flat, i.e. made of equally reliable propositional formulas, the proposal made by Rescher and Manor [21] is very commonly used nowadays: compute the set of maximal consistent subsets of the knowledge base first, then a formula is accepted as a consequence when it can be classically inferred from all maximal consistent subsets of propositions or from at least one maximal consistent subset.

However the first consequence relation is very conservative hence rather unproductive while the latter is too permissive may leads to pairs of mutually exclusive conclusions. A mild inference approach is proposed in this paper, that is more productive than the first consequence relation but do not lead to conclusions which are pairwise contradictory. It is based on the idea of arguments that goes back to Toulmin [25] and is related to previous proposals [19], [18], and [24] that were suggested in the framework of defeasible reasoning for handling exceptions. We suggest that a conclusion can be inferred from an inconsistent knowledge base if the latter contains an argument that supports this conclusion, but no argument that supports its negation.

The paper is organised as follows. Section 2 deals with flat knowledge bases and compares several notions of consequence relations that are inconsistency-tolerant, including several ones that come from the non-monotonic logic literature. Section 3 contains a thorough analysis of our argumentative inference process. Section 4 extends the argumentative inference to layered knowledge bases where layers express degrees of certainty as in possibilistic logic [10]; it refines the flat case by allowing for pieces of information of various levels. Section 5 deals with a paraconsistent-like treatment of layered inconsistent knowledge bases, whereby a formula carries two weights: its degree of certainty and the degree of certainty of its negation. Lastly a new way of combining knowledge bases issued from several sources is suggested, inspired by the argumentative inference. Results are given without proofs due to space limitations. Proofs appear in the full report.

## 2. Arguments in Flat Knowledge bases

### 2.1. Definition of an Argumentative Consequence Relation

For the sake of simplicity, we consider in this paper only a finite propositional language denoted by $\mathcal{L}$. We denote the set of classical interpretations by $\Omega$, by $\vdash$ the classical consequence relation. Let $\Sigma$ be a set of propositional formulas, possibly inconsistent but not deductively closed. $Cn(\Sigma)$ denotes the deductive closure of $\Sigma$, i.e. $Cn(\Sigma) = \{\phi \in \mathcal{L}, \Sigma \vdash \phi\}$. We also assume that the knowledge bases manipulated in this section are flat, which means that all formulas in $\Sigma$ have the same reliability.

**Def.1:** A sub-base $\Sigma_i$ of $\Sigma$ is said to be consistent if it is not possible to deduce a contradiction from $\Sigma_i$, and is said to be maximally consistent if adding any formula $\phi$ from $\Sigma - \Sigma_i$ to $\Sigma_i$ produces the inconsistency of $\Sigma_i \cup \{\phi\}$.

We now introduce the notion of argument:

**Def. 2:** A sub-base $\Sigma_i$ of $\Sigma$ is said to be an argument for a formula $\phi$, if it satisfies the following conditions: (i) $\Sigma_i \nvdash \bot$, (ii) $\Sigma_i \vdash \phi$, and (iii) $\forall \psi \in \Sigma_i, \Sigma_i - \{\psi\} \nvdash \phi$.

Notice that this notion of argument is identical to the one proposed in [24] and is also very similar to the notion of environment used in the terminology of the ATMS [12].

**Def. 3:** A formula $\phi$ is said to be an argumentative consequence of $\Sigma$, denoted by $\Sigma \vdash_{\mathcal{A}} \phi$, if and only if:
(i) there exists an argument for $\phi$ in $\Sigma$, and
(ii) there is no argument for $\neg \phi$ in $\Sigma$.

As a consequence of this definition, if our knowledge base contains only the two contradictory statements $\{\phi, \neg\phi\}$ then the inference $\phi \wedge \neg \phi \vdash_{\mathcal{A}} \psi$ does not hold. In other words, our approach is in agreement with the idea of paraconsistent logics [6], where they reject the principle "ex absurdo quodlibet" which allows the deduction of any formula from an inconsistent base.

It is easy to verify that $\vdash_{\mathcal{A}}$ is non-monotonic. Moreover, if $\Sigma$ is consistent then $\Sigma \vdash \phi$ iff $\Sigma \vdash_{\mathcal{A}} \phi$. It means that the non-monotonicity only appears in the presence of inconsistency, and the argumentative consequence resorts to what Satoh [23] calls "lazy non-monotonic reasoning", an idea also proposed in [15].

## 2.2. Comparative Study of Inconsistency-Tolerant Consequence Relations

In this sub-section we compare our approach to reasoning in the presence of inconsistency to the ones reviewed in [3]. We start this comparative study by presenting the different approaches from the most conservative ones to the most adventurous ones. But first we need some further definitions:

**Def. 4:** A sub-base $\Sigma_i$ of $\Sigma$ is said to be minimal inconsistent if and only if it satisfies the two following requirements: (i) $\Sigma_i \vdash \bot$, and (ii) $\forall \phi \in \Sigma_i, \Sigma_i - \{\phi\} \nvdash \bot$.

From now on, we denote by $Inc(\Sigma)$ the set of propositions belonging to at least one minimal inconsistent sub-base of $\Sigma$, namely:
$Inc(\Sigma) = \{\phi, \exists \Sigma_i \subseteq \Sigma,$ such that $\phi \in \Sigma_i$ and $\Sigma_i$ is minimal inconsistent$\}$

The set $Inc(\Sigma)$ is somewhat related to the "base of nogoods" used in the terminology of the ATMS[12]. Once $Inc(\Sigma)$ is computed, we remove from $\Sigma$ all elements of $Inc(\Sigma)$, the result base is called the free base of $\Sigma$, denoted by $Free(\Sigma)$ [2]. In other words, $Free(\Sigma)$ contains all formulae which are not involved in any inconsistency of the knowledge base $\Sigma$. Now, let us introduce the notion of the Free consequence, denoted by $\vdash_{Free}$:

**Def. 5:** A formula $\phi$ is said to be a free consequence (or a sound consequence) of $\Sigma$, denoted by $\Sigma \vdash_{Free} \phi$, iff $\phi$ is logically entailed from $Free(\Sigma)$, i.e. $\Sigma \vdash_{Free} \phi$ iff $Free(\Sigma) \vdash \phi$

The Free-inference relation is very conservative as we will see later. Let us now recall the approach first proposed in [21]. Let $\Sigma$ be a possibly inconsistent base, $MC(\Sigma)$ be the set of all maximal consistent sub-bases of $\Sigma$. The universal (called also the inevitable) consequence relation is defined in [21] in this way:

**Def. 6:** A formula $\phi$ is said to be a universal consequence or MC-consequence of $\Sigma$, denoted by $\Sigma \vdash_\forall \phi$, iff $\phi$ is entailed from each element of $MC(\Sigma)$, namely:
$\Sigma \vdash_\forall \phi$ iff $\forall \Sigma_i \in MC(\Sigma), \Sigma_i \vdash \phi$

As it has been mentioned above, the Free consequence relation is more conservative than MC-consequence:

**Proposition 1:** Each Free-consequence is also a MC-consequence. The converse is false.

One way of finding the proof of the previous proposition is to notice that:
$$Free(\Sigma) = \bigcap_{\Sigma_i \in MC(\Sigma)} \Sigma_i$$
since if a formula $\phi$ does not belong to $Free(\Sigma)$ then there exists a minimal inconsistent sub-base $\Sigma_k$ containing $\phi$, and therefore there exists at least one maximally consistent sub-base which contains $\Sigma_k$ but not $\phi$, which means that there exists at least one element of $MC(\Sigma)$ which does not contain $\phi$, and consequently $\phi$ does not belong to the intersection of the elements of $MC(\Sigma)$. The converse is also true. Indeed, if $\phi \notin \bigcap_{\Sigma_i \in MC(\Sigma)} \Sigma_i$ then there exists a sub-base $\Sigma_i$ such that $\phi \notin \Sigma_i$, and $\Sigma_i \cup \{\phi\}$ is inconsistent, therefore $\phi$ is not free. Then from the properties of Cn, we find:
$Cn(Free(\Sigma)) = Cn(\bigcap_{\Sigma_i \in MC(\Sigma)} \Sigma_i) \subseteq \bigcap_{\Sigma_i \in MC(\Sigma)} Cn(\Sigma_i)$.

The next propositions compare the MC-consequence to the argumentative consequence:

**Proposition 2:** A formula $\phi$ is an argumentative consequence of $\Sigma$ iff (i) $\exists \Sigma_i \in MC(\Sigma)$, such that $\Sigma_i \vdash \phi$, and (ii) $\not\exists \Sigma_j \in MC(\Sigma)$, such that $\Sigma_j \vdash \neg\phi$.

**Proposition 3:** Each MC-consequence of $\Sigma$ is also an argumentative consequence of $\Sigma$. The converse is false

One of the main drawbacks of MC-consequence is the number of elements in $MC(\Sigma)$ which increases exponentially with the the number of conflicts in the base and in general, it is not possible to take into account all the elements of $MC(\Sigma)$. In [9], it is proposed to select a non-empty subset of $MC(\Sigma)$, denoted by $Lex(\Sigma)$, and computed in the following manner:

$$\Sigma_i \in Lex(\Sigma) \text{ iff } \forall \Sigma_j \in MC(\Sigma), |\Sigma_i| \geq |\Sigma_j|$$

where $|\Sigma|$ is the cardinality of $\Sigma$. This ordering is called the lexicographical ordering, and corresponds to the property of parsimony advocated in diagnostic problems [20]. A probabilistic justification of $Lex(\Sigma)$ can be found in [3].

In order to generate the set of plausible inferences based on $Lex(\Sigma)$ from an inconsistent knowledge base, we use a definition similar to the MC-consequence:

**Def. 7:** A formula $\phi$ is said to be a Lex-consequence of $\Sigma$, denoted by $\Sigma \vdash_{Lex} \phi$, iff it is entailed from each element of $Lex(\Sigma)$, namely:

$$\Sigma \vdash_{Lex} \phi \quad \text{iff} \quad \forall \Sigma_i \in Lex(\Sigma), \Sigma_i \vdash \phi$$

**Proposition 4:** Each MC-consequence of $\Sigma$ is also a Lex-consequence of $\Sigma$. The converse is false.

The Lex-consequence and argumentative consequence are not comparable as we see in the following example:
Example Let $\Sigma = \{A, \neg B \vee \neg A, B, \neg C \vee \neg A, C, \neg A \vee D\}$
We have $Lex(\Sigma) = \{\{\neg B \vee \neg A, B, \neg C \vee \neg A, C, \neg A \vee D\}\}$.
Then $\neg A$ is a Lex-consequence of $\Sigma$ while it is not an argumentative consequence, since A is also present in $\Sigma$. In contrast, D is an argumentative consequence (it derives from $\{A, \neg A \vee D\}$) while it is not a Lex-consequence.

The Lex-consequence may appear as an arbitrary selection from $MC(\Sigma)$ if we consider a semantic point of view. Namely, the following situation may happen: $\Sigma_i \in Lex(\Sigma), \Sigma_j \in MC(\Sigma) - Lex(\Sigma)$ and one may define $\Sigma_k$ logically equivalent to $\Sigma_j$ but $|\Sigma_k| > |\Sigma_i|$. However all introduced consequence relations are syntax-sensitive since $\Sigma$ is not closed. Yet, the counterexample demonstrates that the Lex-consequence may implicitly delete some useful pieces of knowledge (here A). It may result in destroying some arguments, as well as some rebuttals (i.e. formulas whose presence ensure an argument for $\neg\phi$ that inhibits arguments for $\phi$).

Another definition of the consequence relation, called existential relation is also proposed in [21], namely:

**Def. 8:** A formula $\phi$ is said to be an existential consequence of $\Sigma$, denoted by $\Sigma \vdash \exists \phi$, iff there exists at least one element of $MC(\Sigma)$ which entails $\phi$, namely:

$$\Sigma \vdash_\exists \phi \quad \text{iff} \quad \exists \Sigma_i \in MC(\Sigma), \Sigma_i \vdash_\exists \phi$$

It is not hard to see that this approach is the most adventurous one, but unfortunately it has an important drawback, since this approach may lead to inconsistent set of results. Indeed, there may exists $\Sigma_i \vdash \phi$ and $\Sigma_j \vdash \neg\phi$, in which case both $\phi$ and $\neg\phi$ will be deduced.

The following hierarchy summarizes the links existing between the different consequence relations studied here, the edge means the inclusion-set relation between the set of results generated by each inference relation. The top of the diagram thus corresponds to the most conservative inferences. All inferences reduce to the classical one when $\Sigma$ is consistent.

Figure 1: A comparative study of inference relations

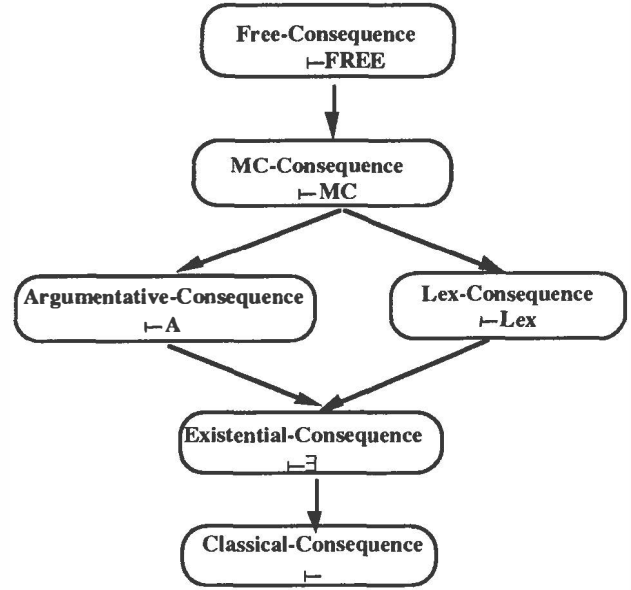

## 3. Properties of $\vdash_\mathcal{A}$:

**Proposition 5** (failure of AND): We may have $\Sigma \vdash_\mathcal{A} \phi$, $\Sigma \vdash_\mathcal{A} \psi$, and *not* $\Sigma \vdash_\mathcal{A} \phi \wedge \psi$.

Proposition 5 must not be seen as a major drawback of $\vdash_\mathcal{A}$ since in some cases we do not want to have the AND property. The $\vdash_\mathcal{A}$ consequence relation captures the cases when we believe in two mutually consistent properties of some object for conflicting reasons.

**Proposition 6:** $\vdash_\mathcal{A}$ satisfies the property of Right Weakening, i.e. If $\phi \vdash \psi$ then $\Sigma \vdash_\mathcal{A} \phi$ implies $\Sigma \vdash_\mathcal{A} \psi$

An important issue when reasoning with an inconsistent knowledge base $\Sigma$ is whether it is possible to construct some equivalent consistent base such that plausible inferences from $\Sigma$ are its logical consequences. In this section we try to construct such an equivalent knowledge base using the argumentative inference relation.

Propositions 5 and 6 are very important to characterise the set of argumentative consequences of a knowledge base $\Sigma$, denoted by $Cn_{\mathcal{A}}(\Sigma)$; $Cn_{\mathcal{A}}(\Sigma)=\{\phi, \Sigma \vdash_{\mathcal{A}} \phi\}$. The fact that the argumentative consequence is not closed under conjunction means that $Cn_{\mathcal{A}}(\Sigma)$ is generally not equal to its closure under Cn, namely: $Cn_{\mathcal{A}}(\Sigma) \neq Cn(Cn_{\mathcal{A}}(\Sigma))$
In this section, we assume that we use only the finite set of propositional symbols *appearing* in the base $\Sigma$.

**Def. 9:** A formula $\phi$ is said to be a prime implicate of $\Sigma$ with respect to the argumentative inference relation if and only if:(i) $\Sigma \vdash_{\mathcal{A}} \phi$, (ii) $\nexists \phi'$, such that $\phi' \vdash \phi$ and $\Sigma \vdash_{\mathcal{A}} \phi'$

A prime implicate can be inferred from a maximal consistent subset of $\Sigma$. However, if $\Sigma_i \in MC(\Sigma)$, then the conjunction of formulas in $\Sigma_i$ (also denoted by $\Sigma_i$) is not a prime implicate since it can be defeated by other maximal consistent subsets of $\Sigma$. Indeed, $\forall\ i \neq j, \Sigma_i \vdash \neg \Sigma_j$. The construction of prime implicates can be achieved from the semantical point view. Indeed, let $[\Sigma_i]$ be the set of models of the maximal consistent sub-base $\Sigma_i$. A given model $\varphi_{\Sigma_i}$ of $[\Sigma_i]$ can be viewed as a formula composed of the conjunction of literals it satisfies. Then it can be shown that the following expression is a prime implicate:

$$\varphi_{\Sigma_1} \vee \cdots \vee \varphi_{\Sigma_{i-1}} \vee \Sigma_i \vee \varphi_{\Sigma_{i+1}} \vee \cdots \vee \varphi_{\Sigma_n}$$

Moreover, if each maximal consistent sub-base is complete (i.e. $\forall a \in \mathcal{L}$, either $a \in \Sigma_i$ or $\neg a \in \Sigma_i$) there exists exactly one prime implicate, and in this case the argumentative consequence and MC-consequence are equivalent. But in general, the prime implicates can be numerous.

Let $R_1,\ldots,R_n$ be the set of prime implicates of $\Sigma$, then $Cn_{\mathcal{A}}(\Sigma)$ can be seen as the union of the deductive closure of each $R_i$ under Cn, namely: $Cn_{\mathcal{A}}(\Sigma)=Cn(R_1)\cup\ldots\cup Cn(R_n)$
And it is easy to check that $\forall\ i, j = 1,n\ \Sigma \nvdash_{\mathcal{A}} R_i \wedge R_j$

Examples

(1) let $\Sigma = \{\neg A \vee B, A \vee C, A, \neg A\}$.
We have two maximal consistent sub-bases,
$\Sigma_1=\{\neg A \vee B, A \vee C, A\}$, $\Sigma_2=\{\neg A \vee B, A \vee C, \neg A\}$
Then:
$[\Sigma_1]=\{A \wedge B \wedge C, A \wedge B \wedge \neg C\}$
$[\Sigma_2]=\{\neg A \wedge B \wedge C, \neg A \wedge \neg B \wedge C\}$,
Therefore we have four prime implicates:
$R_1=B \wedge (C \vee A)$, $R_2=(A \wedge B) \vee (\neg B \wedge C \wedge \neg A)$
$R_3=C \wedge (\neg A \vee B)$, $R_4=(\neg A \wedge C) \vee (\neg C \wedge A \wedge B)$
(2) Consider now $\Sigma$ as:

$\Sigma = \{\neg A \vee B, A \vee B, A, \neg A, C\}$.
We have two maximal consistent sub-bases,
$\Sigma_1=\{\neg A \vee B, A \vee B, A, C\}$, $\Sigma_2=\{\neg A \vee B, A \vee B, \neg A, C\}$
Then:
$[\Sigma_1]=\{A \wedge B \wedge C\}$, $[\Sigma_2]=\{\neg A \wedge B \wedge C\}$
The maximal consistent sub-bases are complete, therefore we have only one prime implicate:
$R=(A \wedge B \wedge C) \vee (\neg A \wedge B \wedge C)=B \wedge C$
Then:
$$Cn_{\mathcal{A}}(\Sigma) = Cn(\{B \wedge C\})$$

Let us now justify why the previous definition makes sense only if we restrict ourselves to the propositional symbols appearing in the knowledge base. Indeed, let us consider the following example $\Sigma=\{A,\neg A\}$, we have only one prime implicate, the tautology T, and therefore it is not possible to deduce $A \vee B$ from $Cn(T)$ which is an argumentative consequence of the knowledge base.

The situation: $\Sigma \vdash_{\mathcal{A}} R_i$, $\Sigma \vdash_{\mathcal{A}} R_j$ and $\Sigma \nvdash_{\mathcal{A}} R_i \wedge R_j$ can correspond to two cases: (i) No argument supporting $R_i \wedge R_j$ can be found. In that case the arguments $\Sigma_i$ and $\Sigma_j$ supporting $R_i$ and $R_j$ respectively are inconsistent. Indeed, if $\Sigma_i \cup \Sigma_j$ is consistent then $\Sigma_i \cup \Sigma_j \vdash R_i \wedge R_j$ and there would exist an argument supporting $R_i \wedge R_j$; (ii) There is an argument for $\neg R_i \vee \neg R_j$.

Anyway the arguments supporting the prime implicates can be viewed as a set of scenarii extracted from $\Sigma$, that express different points of views on what is the actual information contained in $\Sigma$. These points of view are incompatible in the sense that the subsets $\Sigma_i$ and $\Sigma_j$ supporting two prime implicates $R_i$ and $R_j$ should not be mixed (even if not inconsistent). The fact that $Cn_{\mathcal{A}}(\Sigma)$ still reflects conflicts lying in $\Sigma$ can be seen as follows: the argumentative inference forbids that two prime implicates $R_i$ and $R_j$ be inconsistent. However the set $\{R_1, \ldots, R_n\}$ can be globally inconsistent, namely one argumentative consequence of $\Sigma$ can be defeated by other consequences grouped together.

Example
Consider the set $\Sigma=\{\neg A, \neg B, A, B, \neg C \vee \neg D, \neg A \vee B\}$
The maximal consistent subsets of $\Sigma$ are:
$\Sigma_1 = \{\neg A, \neg B, \neg C \vee \neg D, \neg A \vee B\}$
$\Sigma_2 = \{\neg A, B, \neg C \vee \neg D, \neg A \vee B\}$
$\Sigma_3 = \{A, B, \neg C \vee \neg D, \neg A \vee B\}$
$\Sigma_4 = \{\neg B, A, \neg C \vee \neg D\}$
Consider the three formulas:
$\phi_1=(\neg A \wedge \neg B \wedge (\neg C \vee \neg D)) \vee (\neg C \wedge D \wedge (A \vee B))$
$\phi_2=(\neg A \wedge B \wedge (\neg C \vee \neg D)) \vee (C \wedge \neg D \wedge (A \vee \neg B))$
$\phi_3=(A \wedge B \wedge (\neg C \vee \neg D)) \vee (\neg C \wedge \neg D \wedge (\neg A \vee \neg B))$
It is easy to see that $\Sigma_1 \vdash \phi_1$, $\Sigma_2 \vdash \phi_2$ and $\Sigma_3 \vdash \phi_3$, but we never have $\Sigma_i \vdash \neg \phi_j$ for $i \neq j$. Moreover $\phi_1 \wedge \phi_2 \wedge \phi_3 \vdash \bot$.

This result can be viewed as a weakness of the argumentative inference which avoids obvious direct contradictions, but does not escape hidden ones. It confirms the fact that $Cn_{\mathcal{A}}(\Sigma)$ is a heterogeneous set of properties that pertain to distinct views of the world. This means that a question-answering system whereby a question "is it true that $\phi$" is answered by yes or no after computing $\Sigma \vdash_{\mathcal{A}} \phi$ is not really informative enough. The system must also supply the argument for $\phi$. This way of coping with inconsistency looks natural, and the arguments for $\phi$ and $\psi$ should enable the user to decide whether these two plausible conclusions can be accepted together or not.

## 4. Arguments in prioritized knowledge bases

The use of priorities among formulas is very important to appropriately revise inconsistent knowledge bases. For instance, it is proved in [14] that any revision process that satisfies natural requirements is implicitly based on such a set of priorities. Similarly a proper treatment of default rules also leads to prescribe priority levels, e.g.[16]. In these two cases, the handling of priorities has been shown to be completely in agreement with possibilistic logic [10], [2]. Arguments of different levels are also manipulated in [13] in a way completely consistent with possibilistic logic.

In the prioritized case, a knowledge base can be viewed as a layered knowledge base $\Sigma = B_1 \cup ... \cup B_n$, such that formulas in $B_i$ have the same level of priority or certainty and are more reliable than the ones in $B_j$ where $j > i$. This stratification is modelled by attaching a weight $\alpha \in [0,1]$ to each formula with the convention that $(\phi\ \alpha_i) \in B_i, \forall i$ and $\alpha_1 = 1 > \alpha_2 > ... > \alpha_n > 0$.

A sub-base $\Sigma_i = E_1 \cup ... \cup E_n$ of $\Sigma = B_1 \cup ... \cup B_n$ where $\forall j = 1,n, E_j \subseteq B_j$ is said to be consistent if: $\Sigma_i \nvdash \bot$ and is said to be maximal consistent if adding any formula from $(\Sigma - \Sigma_i)$ to $\Sigma_i$ produces an inconsistent knowledge base.

Before introducing the notion of argument in prioritized knowledge base, let us define the notion of entailment in a layered base, named $\pi$-entailment:

**Def. 10:** Let $\Sigma = B_1 \cup ... \cup B_n$ be a layered knowledge base. A formula $\phi$ is said to be a $\pi$-consequence of $\Sigma$ with weight $\alpha_i$, denoted by $\Sigma \vdash_{\pi} (\phi\ \alpha_i)$, if and only if:
(i) $B_1 \cup ... \cup B_i$ is consistent, and
(ii) $B_1 \cup ... \cup B_i \vdash \phi$
(iii) $\forall j < i, B_1 \cup ... \cup B_j \nvdash \phi$

The definition of $\vdash_{\pi}$ is identical to the one proposed in possibilistic logic [7], [8], [10]. It is clear that in the presence of inconsistency the $\pi$-entailment and the classical entailment have not the same behaviour. Indeed in classical logic if our base $\Sigma$ is inconsistent then any formula can be deduced from $\Sigma$ and the base becomes useless. In a stratified base, the situation is better since it is possible to use only a consistent subbase of $\Sigma$ (in general not maximal), denoted by $\pi(\Sigma)$, induced by the levels of priority and defined in this way:
$\pi(\Sigma) = B_1 \cup ... \cup B_i$, such that $\pi(\Sigma)$ is consistent and $B_1 \cup ... \cup B_{i+1}$ is inconsistent

The remaining sub-base $\Sigma - \pi(\Sigma)$ is simply inhibited. It is not hard to check that the following result holds:
$\Sigma \vdash_{\pi} \phi$      iff      $\pi(\Sigma) \vdash \phi$

However, this way of dealing with inconsistency is not entirely satisfactory, since it suffers from a principal drawback named "drowning problem" in [3], as we can see in the following examples:

<u>Examples:</u>
- Let $\Sigma$ be the following stratified knowledge base:
$\Sigma = \{\{\neg A \vee \neg B\}, \{A\}, \{B\}, \{C\}\}$
This notation of the form $\{B_1, B_2, ..., B_n\}$, where the weights are omitted is used for the sake of simplicity. This base is of course inconsistent, and only the subset $\Sigma_i = \{\{\neg A \vee \neg B\}, \{A\}\}$ is kept, and therefore C cannot be deduced despite the fact that C is outside the conflict.
- A particular case of the drowning effect is called "blocking property inheritance" [16]; [2]. This can be illustrated by the following set of stratified defaults:
$\Sigma = \{\{p\}, \{\neg p \vee b, \neg p \vee \neg f\}, \{\neg b \vee f, \neg b \vee w\}\}$
where p, b, f and w means respectively penguin, bird, fly and wings. From this base it is not possible for a penguin to inherit properties of birds (in our example to inherit property of having wings), while the only undesirable property for a penguin is "flying".

One way of solving the drowning problem is to recover the inhibited free defaults, denoted by $IFree(\Sigma)$, and defined in this way:
$IFree(\Sigma) = Free(\Sigma) \cap (\Sigma - \pi(\Sigma))$

Then once the inhibited free set has been computed, we define the new inference relation in this way:

**Def. 11:** A formula $\phi$ is said to be a $\pi$+Free-consequence of $\Sigma$, iff it is logically entailed from $\pi(\Sigma) \cup IFree(\Sigma)$, namely: $\Sigma \vdash_{\pi+free} \phi$ iff $\pi(\Sigma) \cup IFree(\Sigma) \vdash \phi$

**Proposition 7:** Each $\pi$-consequence of $\Sigma$ is also a $\pi$+Free-consequence of $\Sigma$.

Brewka [4] (see also [22]) has proposed a more adventurous approach to reason with inconsistent and layered knowledge bases, the idea is to take advantage of the stratification of the base to rank-order the maximal consistent sub-bases of $\Sigma$ and keep only the best ones, namely the "so-called preferred sub-bases".

Let $\Sigma = B_1 \cup \ldots \cup B_n$ be a layered knowledge base. A preferred sub-base $\Sigma_i$ is constructed by starting with a maximal consistent sub-base of $B_1$, then we add to $\Sigma_i$ as many formulas of $B_2$ as possible (wrt to consistency criterion), and so on. Formally, $\Sigma_i$ is a preferred sub-base of $\Sigma$ if it can be constructed as follows: $\Sigma_i = E_1 \cup E_2 \cup \ldots \cup E_n$, where $\forall j = 1,n, E_1 \cup E_2 \cup \ldots \cup E_j$ is a maximal consistent sub-base of $B_1 \cup B_2 \cup \ldots \cup B_j$.

Preferred subbases have also been independently introduced in [9] in the setting of possibilistic logic under the name of strongly maximal consistent subbases. They are such that $\Sigma_i \cup \{(\phi\ \alpha)\} \vdash_\pi (\bot\ \alpha)$, $\forall (\phi\ \alpha) \in \Sigma - \Sigma_i$.

**Def. 12:** Let $\text{Pref}(\Sigma)$ be the set of preferred sub-bases of $\Sigma$. A formula $\phi$ is said to be a preferred consequence of $\Sigma$, denoted by $\Sigma \vdash_{\text{pref}} \phi$, iff it is entailed from each element of $\text{Pref}(\Sigma)$, namely: $\Sigma \vdash_{\text{pref}} \phi$ iff $\forall \Sigma_i \in \text{Pref}(\Sigma)$, $\Sigma_i \vdash \phi$

**Proposition 8:** Each $\pi$+Free-consequence of $\Sigma$ is also a preferred consequence of $\Sigma$. The converse is false

The Lex-consequence relation described in the case of flat knowledge bases has also been proposed in the case of stratified knowledge bases [9]. The objective is to reduce the number of elements of $\text{Pref}(\Sigma)$, by selecting the elements which satisfy the following requirement:
$\Sigma_i = E_1 \cup \ldots \cup E_n \in \text{Lex}(\Sigma)$ iff $\forall \Sigma_j = E'_1 \cup \ldots \cup E'_n \in \text{Pref}(\Sigma)$,
$\nexists i$, such that $|E'_i| > |E_i|$ and $\forall j < i\ |E'_j| = |E_j|$
The definition of Lex-consequence is identical to the one presented in the case of a flat knowledge base, namely a formula $\phi$ is a Lex-consequence of $\Sigma$ if and only if it is entailed from each element of $\text{Lex}(\Sigma)$.

**Proposition 9:** Each preferred-consequence of $\Sigma$ is also a Lex-consequence of $\Sigma$. The converse is false.

Now, we propose to extend the argumentative inference to layered knowledge bases, and to compare it with the inferences proposed above.

**Def. 13:** A sub-base $\Sigma_i$ of $\Sigma$ is said to be an argument for a formula $\phi$ with a weight $\alpha$ if it satisfies the following conditions: (i) $\Sigma_i \nvdash \bot$, (ii) $\Sigma_i \vdash_\pi (\phi\ \alpha)$, and (iii) $\forall (\psi\ \beta) \in \Sigma_i$, $\Sigma_i - \{(\psi\ \beta)\} \nvdash_\pi (\phi\ \alpha)$

**Def. 14:** A formula $\phi$ is said to be an argumentative consequence of $\Sigma$, denoted by $\Sigma \vdash_{\mathcal{A}} (\phi\ \alpha)$, iff:
(i) there exists an argument for $(\phi\ \alpha)$ in $\Sigma$, and
(ii) for each argument of $(\neg\phi\ \beta)$ in $\Sigma$, we have $\alpha > \beta$.

We now sketch the procedure which determines if $\phi$ is an argumentative consequence of a stratified knowledge base $\Sigma = B_1 \cup \ldots \cup B_n$. The procedure presupposes the existence of an algorithm which checks if there exists an argument for a given formula in some flat base. This can be achieved by using the variant of a refutation method proposed for example in [15].

The procedure is based on a construction of the maximal argument of $\phi$ and its contradiction. First we start with the sub-base $B_1$, and we check if there is a consistent sub-base of $B_1$ which entails $\phi$ or $\neg\phi$. If the response is respectively Yes-No then $\phi$ is an argumentative consequence of $\Sigma$ with a weight $\alpha_1 = 1$, by symmetry if the response is No-Yes then $\neg\phi$ is in this case the argumentative consequence of $\Sigma$. Now if the response is Yes-Yes then neither $\phi$ nor $\neg\phi$ are argumentative consequences. If the response corresponds to one of the answers given above then the algorithm stops. If the response is No-No we repeat the same cycle described above with $B_1 \cup B_2$. The algorithm stops when we have used all the knowledge base $\Sigma$.

As discussed in the case of a flat knowledge base, the inference relation $\vdash_{\mathcal{A}}$ is non-monotonic, and if our knowledge base is consistent then the set of formulas generated by $\vdash_{\mathcal{A}}$ is identical to the one generated by the "possibilistic" inference rule $\vdash_\pi$.

The next proposition shows that $\vdash_{\mathcal{A}}$ is a faithful extension of the inference $\pi$-entailment.

**Proposition 10:** If $\Sigma \vdash_\pi (\phi\ \alpha)$ then $\Sigma \vdash_{\mathcal{A}} (\phi\ \alpha)$. The converse is false.

**Proposition 11:** Each $\pi$+Free-consequence of $\Sigma$ is also an argumentative consequence of $\Sigma$. The converse is false

The argumentative consequence is not comparable to the Pref-consequence nor the Lex-consequence, as we see in the following example:

Example
- Let $\Sigma = \{\{A, \neg B \vee \neg A, B, C\}, \{\neg C \vee \neg A\}, \{\neg A \vee D\}\}$
  We have:
  - $\text{Pref}(\Sigma) = \{\{\{A, \neg B \vee \neg A, C\}, \{\neg A \vee D\}\}, \{\{A, B, C\}, \{\neg A \vee D\}\}, \{\{\neg B \vee \neg A, B, C\}, \{\neg C \vee \neg A\}, \{\neg A \vee D\}\}\}$
  - $\text{Lex}(\Sigma) = \{\{\{\neg B \vee \neg A, B, C\}, \{\neg C \vee \neg A\}, \{\neg A \vee D\}\}\}$.
  Then $\neg A$ is a Lex-consequence of $\Sigma$ while it is not an argumentative consequence, since A is also present in $\Sigma$. Note that one may object to the deletion of A from $\text{Lex}(\Sigma)$, given its high priority. Hence the Lex-consequence looks debatable. In contrast, D is an argumentative consequence (it derives from $\{A, \neg A \vee D\}$ while it is not a Pref-consequence nor a Lex-consequence. Again the Pref-consequence forgets the argument, because A and $\neg A \vee D$ do not belong to all preferred subbases.
- Let $\Sigma = \{\{A\}, \{\neg A\}, \{\neg A \vee \neg D, A \vee D\}\}$, we have $\text{Pref}(\Sigma) = \{\{A\}, \{\neg A \vee \neg D, A \vee D\}\}$. In this case $\neg D$ is a Pref-consequence, while it is not an argumentative consequence of $\Sigma$. Again, the argument for D is killed by $\text{Pref}(\Sigma)$.

As we have done in the non-stratified case, we summarize the relationships between the different consequence relations:

Figure 2: A comparative study of inference relations in stratified knowledge

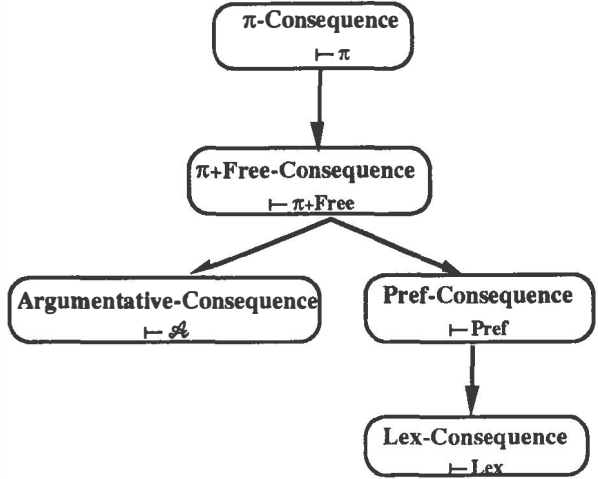

## 5. Paraconsistent-Like Reasoning in Layered Knowledge Bases

In the preceding sections we have seen how in the case of flat and prioritized knowledge bases it is possible to use consistent subparts of it in order to define different types of consequences which are still meaningful. Levels of priority or of certainty attached to formulas have also been used to distinguish between strong and less strong arguments in favor of a proposition or of its contrary. However it is possible to go one step further in the use of the certainty or priority levels by i) attaching to a proposition $\phi$ not only the (greatest) weight $\alpha$ attached to a logical proof of $\phi$ (in the sense of section 4) from a consistent subbase, but also the weight $\beta$ attached to the strongest argument in favor of $\neg\phi$ if any, and ii) by continuing to infer from premises such $(\phi, \alpha, \beta)$ propagating the weights $\alpha$ and $\beta$. It will enable us to distinguish between consequences obtained only from "free" propositions in the knowledge base $\Sigma$ for which $\beta=0$ (i.e. propositions for which there is no argument in $\Sigma$ in favor of their negation), and consequences obtained using also propositions which are not free (for which there exist both a weighted argument in their favor and a weighted argument in favor of their negation).

More formally, the idea is first to attach to any proposition in the considered stratified knowledge base $\Sigma$ two numbers reflecting the extent to which we have some certainty that the proposition is true and to what extent we have some certainty that the proposition is false, and then to provide some extended resolution rule enabling us to infer from such propositions. For each $\phi$, such that $(\phi\ \alpha)$ is in $\Sigma$, we compute the largest weight $\alpha'$ associated with an argument for $\phi$ and the largest weight $\beta'$ associated with an argument for $\neg\phi$ in the sense of Section 4. If there exists no argument in favor of $\neg\phi$, we will take $\beta'=0$; it means in this case that $(\phi\ \alpha)$ is among the free elements of $\Sigma$ since $\phi$ is not involved in the inconsistency of $\Sigma$ (otherwise there would exist an argument in favor of $\neg\phi$).

In the general case, we shall say that $\phi$ has a level of "paraconsistency" equal to $\min(\alpha',\beta')$. Classically and roughly speaking, the idea of paraconsistency, first introduced in [6], is to say that we have a paraconsistent knowledge about $\phi$ if we both want to state $\phi$ and to state $\neg\phi$. It corresponds to the situation where we have conflicting information about $\phi$. In a paraconsistent logic we do no want to have every formula $\psi$ deducible as soon as the knowledge base contains $\phi$ and $\neg\phi$ (as it is the case in classical logic). The idea of paraconsistency is "local" by constrast with the usual view of inconsistency which considers the knowledge base in a global way. It is why we speak here of paraconsistent information when $\min(\alpha',\beta') > 0$. Note that in this process we may improve the lower bound $\alpha$ into a larger one $\alpha'$ if $\exists\ \Sigma_i \subseteq \Sigma$, $\Sigma_i$ consistent and $\Sigma_i \vdash_\pi (\phi\ \alpha')$ (similarly for $\beta'$ if $(\neg\phi\ \beta)$ is already present in $\Sigma$). Then $\Sigma$ is changed into a new knowledge base $\Sigma'$ where each formula $(\phi\ \alpha)$ of $\Sigma$ is replaced by $(\phi\ \alpha'\ \beta')$. Moreover if $\alpha' < \beta'$, i.e. the certainty in favor of $\neg\phi$ is greater than the one in favor of $\phi$, we replace $(\phi\ \alpha'\ \beta')$ by $(\neg\phi\ \beta'\ \alpha')$. If $\phi$ is under a clausal form, $\neg\phi$ is a conjunction $\varphi_1 \wedge ... \wedge \varphi_n$ ; in this case we will replace $(\neg\phi\ \beta'\ \alpha')$ by the clauses $(\varphi_i\ \beta'\ \alpha')$, $i=1,n$ in order to keep $\Sigma'$ under a clausal form if $\Sigma$ was under a clausal form. Let us consider an example

$\Sigma = \{(\neg A \vee B\ \alpha), (A\ \beta)\ (\neg B\ \gamma)\ (B\ \delta), (\neg B \vee C\ \varepsilon), (\neg C\ \rho)\}$.
Then:
$\Sigma' = \{(\neg A \vee B\ \max(\alpha,\delta)\ \min(\beta,\gamma)), (A\ \beta\ \min(\alpha,\gamma)),$
$(\neg B\ \max(\gamma, \min(\rho,\varepsilon)), \max(\delta, \min(\alpha,\beta))),$
$(B\ \max(\delta, \min(\alpha,\beta)), \max(\gamma, \min(\rho,\varepsilon))),$
$(\neg B \vee C\ \varepsilon\ \min(\rho,\delta)), (\neg C\ \rho\ \min(\varepsilon,\delta))\ \}.$

Depending on the ordering between the weights we will keep either $(\neg B\ x\ y)$ or $(B\ y\ x)$ depending if $x > y$ or $y > x$. If $x = y$ we will keep both of them in $\Sigma'$.

In a second step an extended resolution rule can be proposed in order to infer from propositions in $\Sigma'$. This rule expressed in clausal form is (see the full report for a proof, see [8] also):

$$\frac{(A \vee B\ \alpha'\ \beta')\quad (\neg B \vee C\ \gamma'\ \delta')}{(A \vee C\ \varepsilon'\ \rho')}$$

with $\varepsilon' = \min(\max(\gamma',\beta'), \max(\alpha',\delta'))$
$\rho' = \max(\beta',\delta')$.

When $\beta'=\delta'=0$, i.e. the premises are not paraconsistent, we obtain $\varepsilon'=\min(\alpha',\gamma')$, $\rho'=0$. Clearly we have $\varepsilon' \geq \rho'$, i.e. the inference preserves the inequality between the weights. We also observe that the degree of paraconsistency of the conclusion namely $\min(\varepsilon',\rho')=\max(\beta',\delta')$ is equal to the maximum of the degrees of paraconsistency of the two premises namely

min($\alpha'$,$\beta'$)=$\beta'$ and min($\gamma'$,$\delta'$)=$\delta'$. Thus the inference rule extends the standard possibilistic resolution [7, 10] and in case of paraconsistent premise(s), propagates this paraconsistency to the conclusion. In the case where one of the premises is not paraconsistent, i.e. $\beta' = 0$ for instance, the degree of certainty $\varepsilon' = \min(\gamma', \max(\alpha',\delta'))$ of the conclusion is greater than its degree of paraconsistency $\rho' = \delta'$ only if the degree of certainty $\alpha'$ of the non-paraconsistent premise is greater than $\delta'$ and $\gamma'>\delta'$ (i.e. $\gamma' \neq \delta'$). Otherwise the conclusion which is obtained is such that $\varepsilon' = \rho' = \delta'$, i.e. nothing emerges from inconsistency.

Let us consider an example
$\Sigma$ = {(A 1), ($\neg$A$\vee$B  0.8), ($\neg$B 0.6), ($\neg$A$\vee$C  0.5), ($\neg$D 0.3), ($\neg$A$\vee$D  0.4), ($\neg$A$\vee$E  0.7), ($\neg$F$\vee$G  0.5), (F 1), ($\neg$F 0.2), ($\neg$H$\vee$I  0.3), (I 0.4) }.
Observe that $\Sigma \vdash_\pi$ ($\bot$ 0.6), i.e. the global level of inconsistency of the base is 0.6. Then we have
$\Sigma'$ = {(A 1 0.6), ($\neg$A$\vee$B  0,8  0,6), ($\neg$A$\vee$C  0.5  0), (D 0.4  0.3), ($\neg$A$\vee$D  0.4  0.3), ($\neg$A$\vee$E  0.7  0), ($\neg$F$\vee$G  0.5  0), (F 1  0.2), ($\neg$H$\vee$I  0.3  0), (I 0.4  0) }.

Applying the "paraconsistent" resolution rule yields
 (C 0.6  0.6), (E 0.7  0.6), (G 0.5  0.2), (I 0.3  0).
This shows that
–non-paraconsistent premises such as ($\neg$A$\vee$C  0.5  0) with a rather low degree of certainty resolved with another premise (here (A 1 0.6)) whose level of paraconsistency is larger than this degree of certainty, lead to fully blurred paraconsistent conclusions, here (C  0.6  0.6). By contrast $\vdash_{Free+\pi}$ would enable to get (C 0.5), while the refutation procedure used in $\vdash_\pi$ yields (C 0.6), reflecting the global inconsistency of the base
–if the non-paraconsistent premise is sufficiently certain with respect to the paraconsistency of the other premise, e.g. ($\neg$A$\vee$E  0.7 0), and (A 1 0.6), the conclusion, here (E  0.7  0.6) is not completely blurred. This is true even if this certainty is less than the global level of inconsistency of the base (e.g. (G 0.5 0.2) obtained from ($\neg$F$\vee$G  0.5  0), (F 1 0.2))
–if the premises are not paraconsistent, (e.g. ($\neg$H $\vee$ I 0.3 0), (H 0.4 0)), we obtain a non-paraconsistent conclusion, as with $\vdash_{Free+\pi}$, since we do not use refutation.

Generally speaking, if a clause A$\vee$B is more paraconsistent than the clause $\neg$B$\vee$C is certain, then A$\vee$C will be completely blurred by paraconsistency. Indeed from a logical point of view, being more certain that $\neg$A$\wedge\neg$B is true than we are certain that $\neg$B$\vee$C is true, the entailment A$\vee$B, $\neg$A$\wedge\neg$B$\vdash$A$\vee$C applies with a greater level of certainty than A$\vee$B, $\neg$B$\vee$C $\vdash$ A$\vee$C.

We can observe that using the "paraconsistent" resolution rule locally in a knowledge base $\Sigma'$, may yield the same proposition with different weights, namely ($\phi$ $\alpha'$ $\beta'$), ($\phi$ $\alpha''$ $\beta''$). In this case, a more certain and less paraconsistent conclusion should be preferred; this is less obvious when we have to choose between a highly certain but highly paraconsistent conclusion and a conclusion with low certainty and low paraconsistency. Lastly observe that, *for the formulae appearing explicitly in $\Sigma$*, the paraconsistent approach gives the same results as $\vdash_{\mathcal{A}}$. They differ for other conclusions since the paraconsistent approach propagates the effects of local inconsistency.

## 6. Combining knowledge bases

In [1] several approaches are proposed to combine knowledge bases, and one of them is very similar to what [4] calls "preferred sub-theories". The idea is to assume a total ordering between different bases $\Sigma_1>\Sigma_2>... >\Sigma_n$, such that $\Sigma_i$ is more reliable than $\Sigma_j$ for j>i. A resulting base is constructed from $\Sigma_1$ by adding as many formulas as possible from $\Sigma_2$ (wrt consistency criterion), then as many formulas as possible from $\Sigma_3$, and so on. The principal problem is that the resulting base is not unique.

Two approaches are proposed in [5] to merge bases according to suspicious attitude or to trusting attitude. The suspicious attitude is very conservative since for example the result of merging two knowledge bases $\Sigma_1>\Sigma_2$ is equal to the union of the two bases if they are not conflictual, and is equal to $\Sigma_1$ in other cases. In the trusting attitude, the approach is very similar to [1] and produces always one resulting base, but unfortunately the approach is very restrictive since the knowledges bases to be merged must be sets of literals.

In the context of possibilistic logic, an approach has been proposed in [11] for the fusion of n knowledge bases $\Sigma_1,...,\Sigma_n$. One way of defining the resulting base is to consider the intersection of the deductively closed bases $Cn(\Sigma_i), ..., Cn(\Sigma_n)$ (by $\vdash_\pi$). It is clear that this approach is very cautious. In the same paper, another approach has been proposed considering now the union of the deductively closed bases $Cn(\Sigma_i),...,Cn(\Sigma_n)$. However when the resulting base is inconsistent, then some formulas will be inhibited by the drowning effect [3]

We suggest a new approach to merge n knowledge bases $\Sigma_1, ..., \Sigma_n$. For this aim we use of a variation of the argumentative consequence relation, denoted by $\vdash_{\mathcal{A}\mathcal{M}}$, $\mathcal{M}$ for the multi-sources, and which is defined in the following way:
   $\Sigma_1, ..., \Sigma_n \vdash_{\mathcal{A}\mathcal{M}} (\phi$ $\alpha)$ iff:
      $\exists\Sigma_i$, such that $\Sigma_i \vdash (\phi$ $\alpha)$, and
      $\not\exists\Sigma_j$, such that $\Sigma_j \vdash (\neg\phi$ $\beta)$ such that $\beta > \alpha$.
Then the resulting knowledge base is: $\Sigma_{result} = \{(\phi$ $\alpha)$ / $\Sigma_1, ..., \Sigma_n \vdash_{\mathcal{A}\mathcal{M}} (\phi$ $\alpha)\}$. Because $\Sigma_{result}$ can be inconsistent, this approach should be used for question-answering purposes only, and each response should be accompanied with its argument.

## 7. Conclusion

The proposed notion of argumentative inference is appealing for several reasons. First it is an extension of classical inference (in the flat case) and possibilistic inference (in the layered case) that copes with inconsistency in a very "ecological" way. Namely it is very faithful to the actual contents of the knowledge base, and does not do away with information contained in it, as opposed to the approaches based on preferred and lexicographically preferred subbases. It avoids the drowning effect of standard possibilistic logic by salvaging sentences whose level of entrenchment is low but are not involved in any contradiction set. Another advantage is that it is amenable to efficient standard implementation methods based on classical resolution.

Also it avoids outright contradictory responses (such that $\phi$ and $\neg\phi$), although several deduced sentences can be globally inconsistent. But as pointed out earlier, the arguments supporting a set of more than two globally contradictory sentences are distinct, so that the reality of this contradiction is debatable, and only reflects the presence of different points of view. Anyway it seems that it is the price to pay in order to remain faithful to an inconsistent knowledge base. Another result of the paper is the use of local contradictions as a specific weight attached to sentences. This approach only partially avoids the drowning effect, but leads to more informative responses than possibilistic logic since not only the certainty of the formula is evaluated, but also its level of conflict. In the future, the paraconsistent inference should be positioned with respect to the other inference modes in order to assess the benefits of carrying local weights of conflict.

Lastly it would be interesting to apply the above result to default reasoning and compare in such a framework the argumentative inference and the one proposed by [24].